\documentclass[3p,times,twocolumn]{elsarticle}

\journal{Knowledge-Based Systems}

\usepackage[utf8]{inputenc}
\usepackage{times}
\usepackage{url}
\usepackage[hidelinks]{hyperref}
\usepackage[small]{caption}
\usepackage{graphicx}
\usepackage{amsmath,amssymb,amsthm}
\usepackage{booktabs}
\usepackage{algorithm}
\usepackage{algorithmic}
\usepackage{multirow}
\usepackage{multicol}
\usepackage[switch]{lineno}

\biboptions{numbers,sort&compress}

\begin{document}

\begin{frontmatter}

\title{CECOR: Correction-oriented synthetic data construction for factual error correction}

\author[inst1]{Lei Zhu}
\ead{maplesakura@tju.edu.cn}

\author[inst1]{Xiaobao Wang\corref{cor1}}
\ead{wangxiaobao@tju.edu.cn}

\author[inst1]{Jianbiao Yang}

\author[inst2]{Chenyang Wang}

\author[inst1]{Dongxiao He}

\author[inst1]{Longbiao Wang}

\author[inst1]{Jianwu Dang}

\cortext[cor1]{Corresponding author.}

\affiliation[inst1]{
  organization={Tianjin University},
  addressline={135 Yaguan Road, Jinnan District},
  city={Tianjin},
  postcode={300354},
  country={China}
}

\affiliation[inst2]{
  organization={Shenzhen University},
  addressline={3688 Nanhai Avenue, Nanshan District},
  city={Shenzhen},
  postcode={518060},
  country={China}
}

\begin{abstract}
Factual Error Correction (FEC) aims to revise inaccurate text into statements that are factually consistent with external evidence. Although recent methods perform well on single-hop correction, they often treat claims as atomic units and struggle with multi-hop cases that require compositional reasoning across multiple evidence sources. This challenge is further amplified by limited paired data and difficulties in locating semantic errors within complex reasoning chains.
We present CECoR (Compositional Error Correction via Reasoning-aware Synthesis), a reasoning-aware framework that introduces a Decomposition and Injection paradigm for compositional error correction. CECoR decomposes multi-hop claims into interpretable reasoning steps and injects controlled perturbations to synthesize high-quality training pairs. A two-stage learning strategy combining supervised fine-tuning and reinforcement learning improves factual accuracy and robustness.
Comprehensive evaluations show that CECoR achieves strong performance on multi-hop benchmarks, outperforming both distantly supervised methods and few-shot LLM baselines. It also generalizes effectively to single-hop correction and remains stable under noisy evidence, demonstrating its versatility for real-world factual correction.
\end{abstract}


\end{frontmatter}

\section{Introduction}

Factual Error Correction (FEC) has become an increasingly vital task in natural language processing, particularly as large language models demonstrate impressive generative abilities across various domains~\cite{touvron2023llamaopenefficientfoundation}. Yet, their tendency to hallucinate, producing fluent but factually incorrect content, undermines their trustworthiness in real-world deployments. FEC aims to revise such inaccurate claims into factually consistent alternatives grounded in supporting evidence, serving as a key safeguard for reliable AI systems.

\begin{figure}[htbp]
  \centering
  \includegraphics[width=1\linewidth]{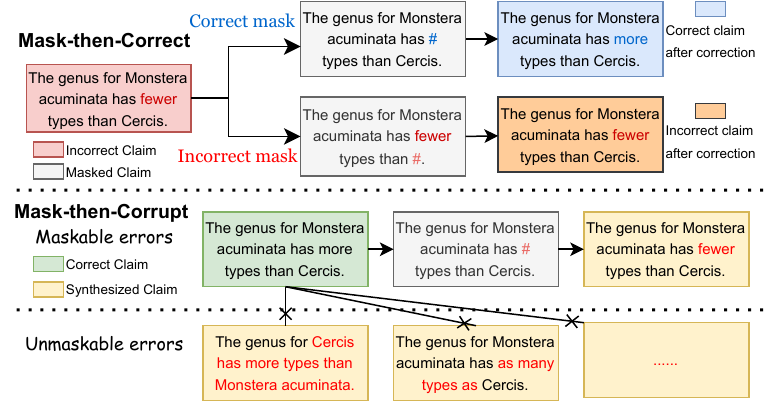}
  \caption{Comparison of factual error correction paradigms.
Top: Mask-then-Correct relies on explicit error masking followed by conditional correction.
Middle: Mask-then-Corrupt generates data by corrupting masked spans but can only handle localized, maskable errors.
Bottom: CECoR enables more diverse and compositional corruptions beyond single-entity substitutions.}
  \label{fig:motivation}
\end{figure}

One widely studied paradigm for FEC is the \textit{Mask-then-Correct} pipeline~\cite{thorne2021evidencebasedfactualerrorcorrection,chen2023convergetruthfactualerror}, illustrated at the top of Figure~\ref{fig:motivation}. This approach decomposes the task into two sequential steps: identifying potential factual errors via masking, and correcting the masked portions conditioned on evidence. While intuitive, it heavily depends on the accuracy of the masking module, which is typically trained under distant supervision due to the lack of large-scale annotated datasets. As a result, failure in the error localization stage often leads to irreparable breakdowns in the entire correction process.

To circumvent this bottleneck, recent methods adopt a reversed formulation known as the \textit{Mask-then-Corrupt} paradigm~\cite{He_Zhang_Jin_Ma_Yuan_Yiu_2024,he-etal-2023-pivotfec}, depicted in the middle of Figure~\ref{fig:motivation}. These methods begin with verified claims, mask factual spans, and then corrupt the masked content to create incorrect variants. This paradigm enables scalable data generation without requiring gold-standard annotations. However, its core limitation lies in its reliance on localized error injection, typically involving the corruption of a single named entity or short span. Such corruption is still predominantly span-local (often limited to a single entity or short phrase), leaving a gap for multi-hop settings where inconsistencies are coupled across multiple reasoning steps and cannot be captured by independent masking operations.

As shown in the bottom portion of Figure~\ref{fig:motivation}, real-world factual errors often exhibit non-local and structurally complex characteristics. The left example illustrates a claim corrupted through syntactic rephrasing that subtly alters temporal logic, while the right example introduces error by jointly replacing multiple entities, affecting relational consistency. These types of factual errors cannot be easily captured by simple masking operations and are common in multi-hop claims, where truth conditions are distributed across reasoning chains involving multiple evidence pieces~\cite{jiang2020hoverdatasetmanyhopfact,aly-etal-2021-fact,yang2018hotpotqadatasetdiverseexplainable}.
More broadly, both Mask-then-Correct and Mask-then-Corrupt rely on an implicit \textit{Atomic Fact Assumption}, which suggests that factual claims are flat, undecomposed units, and that errors are independently correctable. This assumption is not well suited for multi-hop claims that encode compositional truths, where correctness depends on integrating multiple reasoning steps supported by diverse evidence~\cite{pan2023factcheckingcomplexclaimsprogramguided,gao2023palprogramaidedlanguagemodels}. 
Effective correction in such cases requires structure-aware representations and step-wise manipulation of logic, which these paradigms do not provide. Yet, meeting these requirements remains highly non-trivial in practice.

To address these limitations, we propose \textbf{CECoR} (\textbf{C}ompositional \textbf{E}rror \textbf{Co}rrection via \textbf{R}easoning-aware Synthesis), a decomposition-and-injection framework for multi-hop factual error correction. The key idea is to explicitly expose the latent reasoning behind a multi-hop claim by decomposing a verified claim into an interpretable, step-wise reasoning program, and then injecting controlled factual perturbations into one reasoning step at a time before recomposing the corrupted program back into a fluent natural-language claim.
This step-level manipulation enables CECoR to synthesize non-local and coupled inconsistencies, such as coordinated entity substitutions or violations of temporal/relational constraints, that are difficult to express with surface-span masking, while keeping the overall claim coherent. As a result, CECoR converts abundant unpaired correct multi-hop claims into pseudo-parallel (incorrect, correct) pairs, alleviating the scarcity of supervised correction data.
To ensure reliability of the synthesized supervision, we further introduce a filtering pipeline that retains only examples that are fluent, structurally multi-hop, and verifiably inconsistent with the evidence, yielding high-fidelity training pairs for correction.
On top of the synthesized data, we adopt a two-stage learning strategy: (i) supervised fine-tuning on the filtered synthetic pairs to initialize a correction model with compositional error patterns, and (ii) reinforcement learning on naturally occurring incorrect claims to improve robustness and better align corrections with real-world error distributions and editing preferences.
We also revisit evaluation for factual error correction. Existing reference-based, surface-form metrics can over-reward conservative outputs (e.g., leaving the input unchanged) and penalize valid edits, especially when factual discrepancies are subtle. We therefore complement rule-based metrics with an LLM-based evaluation that better reflects semantic correctness and evidence support, following recent findings on LLMs as reliable judges.
Under full evaluation protocol, experiments on multi-hop benchmarks demonstrate strong performance over competitive baselines, with additional generalization to single-hop correction and robustness under noisy retrieved evidence.
Our contributions are summarized as follows:
\begin{itemize}
\item \textbf{Limitation:} We analyze why prior FEC paradigms that operate on undecomposed claims (the atomic fact assumption) are insufficient for multi-hop, compositional inconsistencies.
\item \textbf{Method:} We introduce CECoR, which decomposes multi-hop claims into reasoning programs and performs step-level error injection to synthesize high-fidelity training pairs.
\item \textbf{Learning \& Evaluation: }We propose a two-stage SFT+RL training pipeline and a confidence-oriented evaluation protocol with LLM-based judging, showing strong performance on multi-hop benchmarks and robust transfer to single-hop and noisy evidence settings.
\end{itemize}

\section{Related Work}

\textbf{Factual Error Correction (FEC).}
FEC aims to revise an input claim into a factually supported statement given external evidence, and is closely related to fact verification, grounded generation, and text editing.
A major obstacle in FEC is the scarcity of large-scale paired supervision (incorrect--correct claim pairs) grounded in evidence.
As a result, prior work on complex claims has largely relied on \emph{distant supervision} and \emph{synthetic data construction} to scale training.
Early approaches commonly adopt a ``mask-then-correct'' pipeline, where entities or spans are first masked and a correction model is trained to fill in evidence-consistent content~\cite{thorne2021evidencebasedfactualerrorcorrection,chen2023convergetruthfactualerror}.
This formulation decomposes correction into error localization followed by conditional generation, but its effectiveness is highly dependent on the masking component, which is often trained with weak signals and may fail to localize non-local or implicit errors.

To circumvent explicit localization, subsequent work explores reversed formulations such as ``mask-then-corrupt'', which constructs training pairs by corrupting verified claims after masking selected spans~\cite{he-etal-2023-pivotfec,He_Zhang_Jin_Ma_Yuan_Yiu_2024}.
Methods such as LIFE further improve corruption diversity by learning a corruptor and iteratively refining synthetic examples, enabling scalable supervision without manual annotations~\cite{He_Zhang_Jin_Ma_Yuan_Yiu_2024}.
However, despite these advances, most existing methods share a fundamental limitation: they treat a claim as a flat text sequence and primarily inject \emph{span-local} perturbations.
This implicitly assumes that factuality can be corrected by independently editing a small surface span (often a single entity), which is inadequate for multi-hop settings where inconsistencies can be coupled across multiple reasoning steps, involve relational or temporal constraints, and require compositional alignment between evidence and intermediate sub-facts.

\textbf{Multi-hop reasoning and fact verification.}
Our work is also related to multi-hop fact verification benchmarks and reasoning-centric modeling.
Datasets such as HOVER and FEVEROUS provide claims supported or refuted by multiple evidence pieces, motivating models that can aggregate evidence and perform compositional reasoning rather than relying on single-sentence matching~\cite{jiang2020hoverdatasetmanyhopfact,aly-etal-2021-fact}.
Related work in multi-hop QA and verification highlights that exposing intermediate reasoning signals can reduce error propagation and improve robustness, especially when evidence is distributed across documents or requires chaining~\cite{yang2018hotpotqadatasetdiverseexplainable}.
These observations suggest that multi-hop FEC requires not only stronger generators, but also structure-aware representations that make multi-step dependencies explicit.

\textbf{Program-guided reasoning and decomposition.}
To address the lack of explicit structure in claim-level correction, we draw inspiration from recent advances in program-guided reasoning, which have shown success in fact \emph{verification} by decomposing complex claims into structured, interpretable programs~\cite{pan2023factcheckingcomplexclaimsprogramguided}.
By executing step-wise reasoning over atomic facts, program-based approaches improve interpretability and provide tighter evidence alignment than monolithic sequence models.
In parallel, decomposition strategies in complex reasoning (e.g., least-to-most prompting or reasoning-trace methods) further support the idea that breaking down a hard task into simpler subproblems improves generalization and reduces cascading failures.
We extend this line of work from verification to the more challenging \emph{correction} setting, which requires not only identifying where a reasoning chain breaks but also generating a fluent and factually consistent revision that preserves as much original content as possible.

\textbf{Synthetic data, counterfactual perturbations, and evaluation.}
Our framework is additionally connected to data synthesis and counterfactual learning, where minimal, targeted perturbations are used to create informative training pairs and mitigate reliance on spurious surface cues.
Step-level perturbations naturally produce counterfactual variants that flip factuality while keeping unrelated semantics unchanged, which is beneficial for learning grounded correction behavior.
Finally, recent work has shown that rule-based metrics may misrepresent semantic quality for generation and editing tasks, motivating LLM-based evaluation as a proxy for human judgments in settings where references are incomplete or multiple valid rewrites exist~\cite{zheng2023judgingllmasajudgemtbenchchatbot}.
This motivates our evaluation protocol and our reinforcement learning stage that explicitly optimizes for evidence-grounded semantic correctness.

\textbf{Positioning.}
In summary, prior FEC methods rely heavily on span-level masking/corruption under an implicit atomic-fact assumption, limiting their ability to model multi-step dependencies in multi-hop claims.
Our work is the first to leverage program-style logical decomposition for FEC and to perform \emph{reasoning-step-level} error injection for scalable supervision.
By explicitly modeling intermediate steps, our approach enables higher-quality synthetic data that better reflects realistic error patterns and yields correction models that are more interpretable, generalizable, and robust for complex factual revision.

\begin{figure*}[htbp]
  \centering
  \includegraphics[width=1\linewidth]{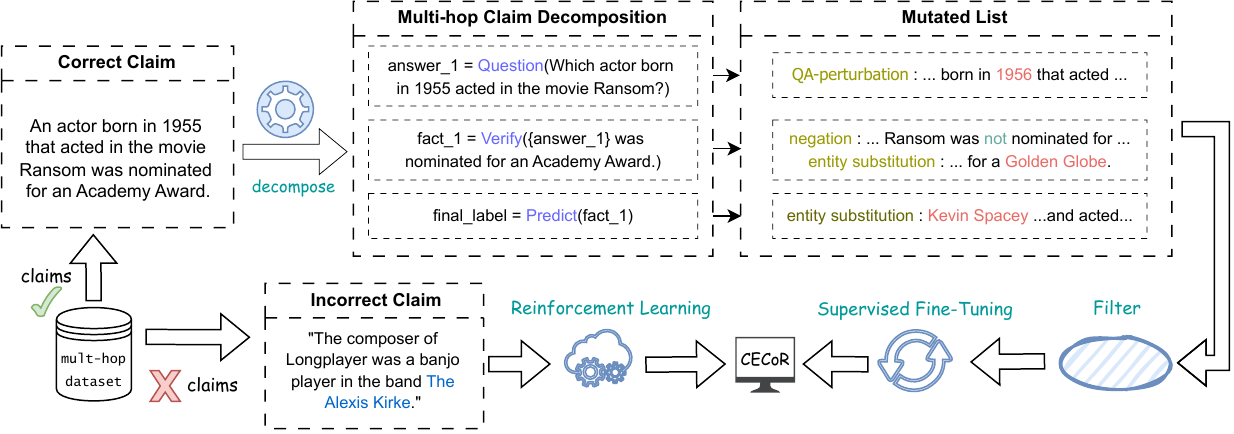}
  \caption{Given multi-hop correct claims, we decompose them into logical steps, inject structured errors, and apply filtering to generate high-quality synthetic data for supervised fine-tuning. In parallel, naturally incorrect claims are used to optimize the model via reinforcement learning. This unified pipeline enables the effective use of both correct and incorrect claims for multi-hop factual error correction.}
  \label{fig:method}
\end{figure*}

\section{The \textsc{CECoR} Framework}
\subsection{Problem Definition}
We define a multi-hop incorrect claim $M$ as a sentence composed of single-hop claims ($C_1, \dots, C_n$) that is factually inconsistent with a given evidence $E$. 
The goal of correction is to revise $M$ into a new claim $M'$ that:
(1) is grammatical and fluent,
(2) is fully supported by the evidence $E$,
(3) resolves all factual inconsistencies present in $M$, and
(4) preserves as much of the original content as possible.

\subsection{Framework Overview}
We propose \textsc{CECoR}, a reasoning-aware framework that combines synthetic data generation and two-stage training, i.e., Supervised Fine-Tuning (SFT) and Reinforcement Learning (RL), to correct complex multi-hop factual claims. Unlike prior claim-level FEC methods, \textsc{CECoR} explicitly models the internal reasoning structure of claims and injects structured errors at the logical step level.

Importantly, this approach enables effective multi-hop factual error correction \textit{without requiring large-scale manually paired data}. By generating diverse, high-fidelity training instances from unpaired correct claims, \textsc{CECoR} substantially lowers the annotation burden while expanding the correction coverage. Furthermore, a model trained under this framework exhibits enhanced \textit{generalization ability} and a more \textit{comprehensive correction scope}, capable of handling a wider variety of complex, real-world factual errors.

As shown in the Figure~\ref{fig:method}, the framework consists of five components: (1) \textbf{Multi-hop Claim Decomposition}: Decompose correct claims into reasoning chains. (2) \textbf{Reasoning-aware Error Injection}: Inject factual errors into logic steps and recompose faulty claims. 
(3) \textbf{Filtering}: Ensure the fluency, correctness, and multi-hop integrity of synthetic data.
(4) \textbf{Supervised  Learning}: Train a correction model using high-quality synthetic pairs.
(5) \textbf{Reinforcement Learning with Natural Claims}: Improve robustness using incorrect claims with RL.

Compared to claim-level masking, program decomposition provides an explicit alignment between evidence-grounded sub-facts and editable reasoning units, which reduces the ambiguity of where to edit under distant supervision, and allows us to control the \emph{type} and \emph{scope} of injected inconsistencies.

Starting with correct claims $\mathcal{M}_o = \{M_o^1, \ldots, M_o^n\}$, we decompose each $M_o^i$ into logical chains, inject errors, and generate faulty versions $\mathcal{M}^i_f = \{M_f^{i1}, \ldots, M_f^{im}\}$. Subsequently, we leverage incorrect claims from datasets in an RL stage to further enhance the model's robustness and generalization to diverse factual error types.

\subsection{Multi-hop Claim Decomposition}
To explicitly expose the underlying reasoning structure of multi-hop claims, we adopt a program-guided decomposition strategy inspired by \textsc{ProgramFC}~\cite{pan2023factcheckingcomplexclaimsprogramguided}. This choice is motivated by prior evidence that program-guided decomposition improves verification of complex, multi-hop claims by making intermediate sub-tasks explicit and executable, thereby strengthening both interpretability and reasoning reliability. For each correct claim $M_o^i$, a planner generates a reasoning program $\mathcal{P} = [S_1, S_2, \ldots, S_k]$, where each step $S_j$ is written in controlled natural language and falls into one of three functional categories: 1) \texttt{QUESTION}: Queries used to extract atomic facts from evidence; 2) \texttt{VERIFY}: Binary checks against retrieved facts; 3) \texttt{PREDICT}: Logical compositions of prior steps, forming conclusions:
\begin{equation}
    \begin{split}
    \mathcal{P} &= \mathrm{Planner}(M_o^i)
    = [S_1,\ldots,S_k], \\
    S_i &\in \{\texttt{QUESTION},\,\texttt{VERIFY},\,\texttt{PREDICT}\}.
    \end{split}
    \label{eq:program_decomposition}
\end{equation}
This transformation turns a linguistically opaque multi-hop claim into an interpretable, compositional reasoning chain. It provides the structural substrate for injecting errors in a semantically meaningful and controllable manner.

\subsection{Reasoning-aware Error Injection}
Given a decomposed program $\mathcal{P} = [S_1, \dots, S_k]$, we introduce factual errors at the level of individual reasoning steps to construct diverse faulty claims. To ensure the semantic coherence of the recomposed claim and avoid cascading inconsistencies, we inject errors into only one step $S_j$ at a time. This design not only prevents uncontrolled interaction effects between multiple corrupted steps but also increases the diversity of resulting training examples. For a \texttt{PREDICT} step (e.g., \texttt{label = Predict(fact$_1$ and fact$_2$ ...)}), we identify a shared entity $e$ across the facts, substitute it with a semantically plausible but incorrect alternative $e'$, and rewrite the modified facts into a single coherent sentence. For a \texttt{VERIFY} step (e.g., \texttt{fact$_i$ = Verify("...")}), we extract a salient factual unit $f$ and either replace it with a related incorrect variant $f'$ or rewrite the statement in its negated form to create a contradiction. For a \texttt{QUESTION} step (e.g., \texttt{ans = Question("Q")}), we identify the correct answer $a$, substitute it with a plausible incorrect answer $b$, construct a modified question $Q'$ for which $b$ is valid, and finally generate a fluent declarative sentence combining $Q'$ and $b$:
\begin{equation}
\begingroup\footnotesize
\setlength{\jot}{1pt}
\begin{gathered}
\tilde{S}_j=
\begin{cases}
S_j\!\left[e\leftarrow e'\right], & S_j\in\texttt{PREDICT},\\
S_j\!\left[f\leftarrow f'\right]\ \text{and}\ \neg S_j, & S_j\in\texttt{VERIFY},\\
S_j\!\left[a\leftarrow b,\ Q\leftarrow Q'\right], & S_j\in\texttt{QUESTION}.
\end{cases}
\\
\mathcal{M}_f^{im}
= \mathrm{Recompose}\!\left([S_1,\ldots,\tilde{S}_j,\ldots,S_k]\right),\\
\end{gathered}
\endgroup
\label{eq:step_level_injection}
\end{equation}
By applying these module-specific transformations to $\tilde{S}_j$ and recomposing the program into natural language, we derive a structurally grounded and semantically diverse set of erroneous multi-hop claims $\mathcal{M}^i_f$ from each correct claim $M_o^i$.

\subsection{Filtering}
Although the reasoning-aware injection process produces structurally valid candidates, the generated claims may still suffer from issues such as incoherence, redundancy, or failed corruption. To ensure training reliability, we apply a multi-step filtering procedure. Given a generated pair $(M_o^i, M_f^{ij})$, we retain it only if it satisfies five criteria: (1) the length of $M_f^{ij}$ falls within a valid range $[l_{\text{min}}, l_{\text{max}}]$, (2) $M_f^{ij} \neq M_o^i$ (identity check), (3) $M_f^{ij}$ maintains multi-hop dependencies (i.e., requires at least two reasoning steps), (4) $M_f^{ij}$ is grammatically fluent based on perplexity and language model scoring, and (5) a factual inconsistency is confirmed via contradiction against the original evidence $E$. Let $\mathcal{F}$ denote this filtering function, then the retained dataset is defined as 
\begin{equation}
    \mathcal{D}_{\text{syn}} = \{(M_f^{ij}, M_o^i) \mid \mathcal{F}(M_f^{ij}, M_o^i) = 1\},
\end{equation}
\begin{equation}
 \mathcal{F}(M_f^{ij}, M_o^i) = 
\begin{cases}
1, & \text{if } M_f^{ij} \text{ passes all filters}, \\
0, & \text{otherwise}.
\end{cases}   
\end{equation}

\subsection{Supervised Learning}
Using the filtered dataset $\mathcal{D}_{\text{syn}}$, we perform supervised fine-tuning (SFT) to train a correction model $\mathcal{M}_{\text{SFT}}$. Let $\theta$ denote model parameters and $\mathcal{L}_{\text{SFT}}$ the training loss. Given a training pair $(M_f^{ij}, M_o^i)$, the objective is to minimize the negative log-likelihood:  
\begin{equation}
\mathcal{L}_{\text{SFT}}(\theta) = -\sum_{(M_f^{ij}, M_o^i)\in\mathcal{D}_{\text{syn}}}\log P_{\theta}(M_o^i \mid M_f^{ij}).
\end{equation}
This trains the model to rewrite faulty multi-hop claims into correct ones using only weakly aligned synthetic supervision. Since no paired gold annotations are required, this approach scales with synthetic generation and provides a strong factual initialization.

\subsection{Reinforcement Learning with Natural Claims}
To enhance the robustness of the correction model against real-world factual inconsistencies beyond synthetic distributions, we introduce a reinforcement learning (RL) stage utilizing incorrect claims \( \mathcal{M}_{ref} = \{M_{ref}^1, \dots, M_{ref}^l\} \) sampled from the \texttt{REFUTES} subset of fact verification datasets. For each incorrect claim \( M_{ref}^x \in \mathcal{M}_{ref} \), the model learns to generate a corrected version \( M_{cor}^x \) that maximizes the expected reward:
$R(M_{ref}^x, M_{cor}^x) = 
\mathcal{S}_{correct}(M_{cor}^x, E) +  
\mathcal{S}_{sim}(M_{ref}^x, M_{cor}^x) + 
\mathcal{S}_{flu}(M_{cor}^x)$
where \( \mathcal{S}_{correct} \) evaluates factual consistency against the evidence \( E \), \( \mathcal{S}_{sim} \) encourages semantic similarity to the original incorrect claim \( M_{ref}^x \), and \( \mathcal{S}_{flu} \) assesses grammaticality and fluency. This yields our final \textbf{CECoR} model, which generalizes effectively to both synthetic and naturally occurring factual errors.

\begin{table*}[htbp]
\centering
\begin{tabular}{p{3cm}ccccc|ccccc}
\toprule
& \multicolumn{5}{c}{\textbf{HOVER}} & \multicolumn{5}{c}{\textbf{FEVEROUS}} \\
\cmidrule(lr){2-6} \cmidrule(lr){7-11}
\multicolumn{1}{c}{{\Large\textbf{Models}}} & \multicolumn{4}{c}{\textbf{SARI-score}} & & \multicolumn{4}{c}{\textbf{SARI-score}} & \\
\cmidrule(lr){2-5} \cmidrule(lr){7-10}
& \textbf{Keep} & \textbf{Add} & \textbf{Delete} & \textbf{Final} & \textbf{RG-2} & \textbf{Keep} & \textbf{Add} & \textbf{Delete} & \textbf{Final} & \textbf{RG-2}\\
\midrule

\multicolumn{11}{l}{\small\textbf{Distantly Supervised Baselines}} \\
\quad LIFE\textsubscript{t5} & 75.66 & 4.27 & 55.46 & 45.13 & 0.72 & 87.40 & 31.59 & 65.35 & 61.45 & 0.81 \\
\quad VENCE\textsubscript{t5} & 81.64 & 2.96 & 73.71 & 52.77 & 0.80 & 62.17 & 5.03 & 80.79 & 49.33 & 0.56 \\

\midrule
\multicolumn{11}{l}{\small\textbf{Few-shot Baselines}} \\
\quad GPT\textsubscript{4o-0shot} & 78.23 & 24.45 & 53.81 & 52.16 & 0.72 & 88.00 & 39.32 & 67.71 & 65.01 & 0.81 \\
\quad GPT\textsubscript{4o-1shot} & 82.99 & 29.43 & 61.58 & 58.00 & 0.79 & 87.43 & 36.45 & 62.95 & 62.28 & 0.81 \\
\quad GPT\textsubscript{4o-4shot} & 83.99 & 28.97 & 62.36 & 58.44 & 0.79 & 90.50 & 43.50 & 75.52 & 69.84 & 0.85 \\
\quad GPT\textsubscript{4o-8shot} & 84.02 & 28.78 & 62.86 & 58.55 & 0.79 & 90.45 & 42.60 & 73.84 & 68.96 & 0.86 \\
\quad LLaMA3-1b & 42.03 & 3.56 & 34.01 & 26.53 & 0.34 & 75.56 & 8.74 & 63.58 & 49.29 & 0.69 \\
\quad LLaMA3-3b & 63.16 & 10.14 & 45.58 & 39.63 & 0.54 & 51.91 & 7.34 & 46.91 & 35.39 & 0.46 \\
\midrule
\multicolumn{11}{l}{\small\textbf{Do-Nothing Baseline}} \\
\quad Do-Nothing & 84.62 & 4.22 & 100 & 62.95 & 0.83 & 88.17 & 3.61 & 100 & 63.93 & 0.85 \\
\midrule
\multicolumn{11}{l}{\small\textbf{Our Methods}} \\
\quad CECoR\textsubscript{T5-sft} & \textbf{91.18} & \textbf{49.35} & \textbf{89.76} & \textbf{76.76} & \textbf{0.87} & \textbf{94.39} & \textbf{54.18} & \textbf{89.48} & \textbf{79.35} & \textbf{0.90} \\
\quad CECoR\textsubscript{L3-1b-sft} & 89.94 & 43.27 & 85.30 & 72.83 & 0.84 & 92.40 & 45.31 & 81.42 & 73.04 & 0.84 \\
\quad CECoR\textsubscript{L3-3b-sft} & \textbf{91.18} & \underline{48.75} & \underline{85.57} & \underline{75.18} & \underline{0.86} & \underline{93.63} & \underline{47.78} & \underline{82.17} & \underline{74.53} & \underline{0.85} \\

\bottomrule
\end{tabular}
\caption{Comparison on the HOVER and FEVEROUS multi-hop factual error correction datasets.}
\label{tab:main_results}
\end{table*}

\section{Experiments}
\subsection{Experimental Setups}
\textbf{Datasets.}
To assess the effectiveness of our framework in correcting complex factual errors requiring multi-hop reasoning, we conduct experiments on two benchmark datasets: \textbf{HOVER}~\cite{jiang2020hoverdatasetmanyhopfact} and \textbf{FEVEROUS}~\cite{aly-etal-2021-fact}. HOVER consists of claims accompanied by multiple supporting evidence sentences from Wikipedia, with each instance labeled as \texttt{SUPPORTS} or \texttt{REFUTES}. Since HOVER lacks paired incorrect/correct claims, we generate a test set by applying our error injection method to supported examples in the validation set as the official test set is unavailable.
To evaluate broader applicability, we also test on \textbf{FEVEROUS}, a multi-hop fact-checking dataset that integrates textual and tabular Wikipedia evidence. As our focus is textual correction, we select only claims that are verifiable using sentence-level evidence. Like HOVER, FEVEROUS does not provide correction pairs, making it well-suited to our distant supervision setting.
To validate generalization beyond multi-hop reasoning, we further evaluate on \textbf{FECDATA}~\cite{thorne2021evidencebasedfactualerrorcorrection}, a manually curated single-hop factual error correction dataset derived from FEVER~\cite{thorne-etal-2018-fever}. It provides high-quality, evidence-grounded incorrect/correct claim pairs. Claims from the \texttt{REFUTED} subset test models’ ability to fix errors, while the \texttt{SUPPORTED} subset ensures factual preservation. The setting serves as a strong testbed for evaluating robustness to simpler correction tasks.Summary statistics of all datasets are reported in the Appendix 1.

\textbf{Baselines.}
We compare our proposed method, \textbf{CECoR}, against nine baselines spanning three categories.
Under the \textit{distant supervision} setting, \textbf{LIFE}~\cite{He_Zhang_Jin_Ma_Yuan_Yiu_2024} learns an explicit corruptor to synthesize faulty input claims for supervised correction training, while \textbf{VENCE}~\cite{chen2023convergetruthfactualerror} applies an iterative \emph{mask-then-correct} procedure until a factually supported output is produced or a maximum iteration limit is reached. Both methods use \textbf{T5-base} as the backbone architecture. Under the \textit{few-shot prompting} setting, we evaluate two lightweight open-source LLMs, \textbf{LLaMA3-1B} and \textbf{LLaMA3-3B}~\cite{grattafiori2024llama3herdmodels}, each prompted with \textbf{8-shot} in-context setting as did in ~\cite{He_Zhang_Jin_Ma_Yuan_Yiu_2024}. In addition, to better assess the capabilities of a stronger proprietary model, we include \textbf{GPT\textsubscript{4o-mini}}~\cite{openai2024gpt4technicalreport} under four in-context learning configurations: \textbf{1-shot}, \textbf{2-shot}, \textbf{4-shot}, and \textbf{8-shot} (The following text is uniformly referred to as \textsubscript{GPT-4o-xshot}), yielding four separate baselines. To ensure a fair comparison across prompting baselines, we adopt the same instruction template and input formatting for all prompted models, and use the same demonstration pool and sampling strategy for selecting examples. We also include a \textit{do-nothing} baseline that outputs the input claim unchanged, serving as a lower bound for correction effectiveness.

\textbf{CECoR Configurations.}
Due to the longer context requirements of multi-hop correction, we adopt \textbf{LLaMA3-1B} and \textbf{LLaMA3-3B} with 4096-token window as our base model. For fair comparison with LIFE and VENCE, we train five variants of our method: \textbf{CECoR\textsubscript{T5-sft}}: T5-base with supervised fine-tuning (SFT), \textbf{CECoR\textsubscript{L3-1b-sft}}, \textbf{CECoR\textsubscript{L3-3b-sft}}: LLaMA3-1B and LLaMA3-3B with SFT, \textbf{CECoR\textsubscript{L3-1b-rl}}, \textbf{CECoR\textsubscript{L3-3b-rl}}: LLaMA3-1B and LLaMA3-3B with SFT and RL. Following common practices, we discard examples that exceed the model’s context length for T5 to ensure fair capacity usage during training and inference. The relevant hyperparameters for model training can be found in the Appendix 2. Since our goal is to better evaluate the effectiveness of the overall framework, rather than to focus on optimizing a single high-performing module, we adopt GPT-4o mini as the concrete instantiation and reward model within our framework after balancing performance and efficiency. The detailed prompts used for LLM calls in each stage are provided in the Appendix 5.

\noindent\textbf{Evaluation Metrics.} 
For automatic evaluation, we use the SARI~\cite{Xu-EtAl:2016:TACL} and ROUGE-2~\cite{lin-2004-rouge} metrics. SARI is decomposed into \textbf{Keep}, \textbf{Delete}, and \textbf{Add} components, with their average representing the final score. ROUGE-2 measures bigram overlap between the revised and reference claim. However, these metrics may reward unchanged inputs or penalize valid paraphrases. To mitigate this limitation, we also employ an LLM-based evaluation framework. \textbf{GPT} denotes GPT-4o-mini, \textbf{DS} denotes DeepSeek-V3, and \textbf{Ge} denotes Gemini-2.5-flash. This method avoids reference bias and better reflects semantic correctness, aligning with recent work on using LLMs as evaluators~\cite{zheng2023judgingllmasajudgemtbenchchatbot}. \textbf{Bold} and \underline{underline} indicate the best and second-best results, respectively, 
excluding the do-nothing baseline.

\subsection{Experimental Results}
\textbf{Main Results.}
Table~\ref{tab:main_results} presents the experimental results on the HOVER and FEVEROUS. The findings reveal clear distinctions among distantly supervised models, few-shot LLM baselines, and our proposed method. Distantly supervised baselines such as LIFE and VENCE, originally designed for simpler single-hop settings, struggle with the compositional semantics of multi-hop correction. On HOVER, these models show relatively low performance across all rule-based metrics. While LIFE performs better on the structurally simpler FEVEROUS dataset, its performance drops significantly on HOVER, suggesting its corruptor module fails to introduce meaningful yet learnable errors for complex claims. VENCE, in contrast, demonstrates better generalization on HOVER, indicating the advantage of its iterative correction mechanism in handling longer, reasoning-heavy claims.
Few-shot prompting with large language models produces strong results. GPT\textsubscript{4o-mini} significantly outperform distantly supervised methods across most metrics, highlighting the capability of instruction-tuned LLMs to generalize with minimal supervision, particularly on less challenging inputs. However, performance on HOVER is relatively weaker, suggesting that few-shot prompting may be insufficient to address the nuanced logical dependencies of multi-hop claims without structured modeling.
Our method achieves state-of-the-art performance across both benchmarks, despite being built on relatively lightweight backbones. The CECoR\textsubscript{T5-sft} model obtains the highest SARI and RG-2 scores on both datasets, while CECoR\textsubscript{LLaMA3-3b-sft} follows closely, confirming the utility of our decomposition and error injection framework in generating high-quality synthetic supervision.

\begin{table}[htbp]
\renewcommand{\arraystretch}{1}
\setlength{\tabcolsep}{1.5mm}
\centering
\begin{tabular}{lccc|ccc}
\toprule
& \multicolumn{3}{c}{\textbf{HOVER}} & \multicolumn{3}{c}{\textbf{FEVEROUS}} \\
\cmidrule(lr){2-4} \cmidrule(lr){5-7}
\multicolumn{1}{c}{{\Large\textbf{Models}}} & \textbf{GPT} & \textbf{DS} & \textbf{Ge} & \textbf{GPT} & \textbf{DS} & \textbf{Ge}\\
\midrule

\multicolumn{7}{l}{\small\textbf{Distantly Supervised Baselines}} \\
\quad LIFE\textsubscript{t5} & 0.23 & 0.26 & 0.19 & 0.66 & 0.72 & 0.71 \\
\quad VENCE\textsubscript{t5} & 0.11 & 0.10 & 0.10 & 0.15 & 0.21 & 0.06 \\

\midrule
\multicolumn{7}{l}{\small\textbf{Few-shot Baselines}} \\
\quad GPT\textsubscript{4o-0shot} & 0.75 & \textbf{0.80} & \textbf{0.85} & \underline{0.85} & \textbf{0.88} & \textbf{0.96} \\
\quad GPT\textsubscript{4o-1shot} & 0.64 & 0.74 & 0.79 & 0.84 & \underline{0.87} & \underline{0.94} \\
\quad GPT\textsubscript{4o-4shot} & \underline{0.68} & 0.76 & \underline{0.81} & 0.81 & 0.85 & 0.93 \\
\quad GPT\textsubscript{4o-8shot} & 0.65 & 0.75 & \underline{0.81} & 0.81 & 0.85 & 0.93 \\
\quad LLaMA3-1b & 0.22 & 0.23 & 0.18 & 0.32 & 0.36 & 0.26 \\
\quad LLaMA3-3b & 0.37 & 0.39 & 0.35 & 0.21 & 0.24 & 0.16 \\
\midrule
\multicolumn{7}{l}{\small\textbf{Do-Nothing Baseline}} \\
\quad Do-Nothing & 0.05 & 0.03 & 0.09 & 0.11 & 0.18 & 0.07 \\
\midrule
\multicolumn{7}{l}{\small\textbf{Our Methods}} \\
\quad CECoR\textsubscript{T5-sft} & 0.43 & 0.61 & 0.68 & 0.67 & 0.79 & 0.83 \\
\quad CECoR\textsubscript{L3-1b-sft} & 0.34 & 0.55 & 0.60 & 0.62 & 0.74 & 0.74 \\
\quad CECoR\textsubscript{L3-3b-sft} & 0.45 & 0.59 & 0.68 & 0.69 & 0.79 & 0.81 \\
\quad CECoR\textsubscript{L3-1b-rl} & 0.62 & 0.62 & 0.58 & 0.74 & 0.83 & 0.82 \\
\quad CECoR\textsubscript{L3-3b-rl} & \textbf{0.83} & \textbf{0.80} & 0.80 & \textbf{0.87} & \underline{0.87} & 0.92 \\

\bottomrule
\end{tabular}
\caption{Comparison on the HOVER and FEVEROUS under LLM-based evaluation. 
}
\label{tab:llm}
\end{table}
\textbf{LLM as a Judge.}
The Do-Nothing baseline, despite performing no correction, achieves unexpectedly high SARI and RG-2 scores, occasionally surpassing competitive systems. This exposes a key weakness of surface-form metrics: their dependence on lexical overlap can yield misleadingly high scores even without factual improvement.
To better capture semantic correctness, we adopt an LLM-based evaluation framework~\cite{zheng2023judgingllmasajudgemtbenchchatbot}, as reported in Table~\ref{tab:llm}. Under this protocol, the Do-Nothing baseline receives appropriately low ratings, demonstrating the discriminative capability of LLM judges and validating our evaluation setup. Our RL-enhanced models—particularly CECoR\textsubscript{L3-3b-rl}—achieve the best GPT-based scores, outperforming strong LLM baselines and confirming the effectiveness of our approach.
Scores from multiple independent LLM judges show consistent trends, indicating the robustness of this semantic evaluation. In contrast, distantly supervised receive much lower GPT-based scores, suggesting that their outputs remain semantically inadequate when judged by strong LLM evaluators. Although the RL-enhanced variants show slightly lower rule-based scores, this is expected, as high-quality factual corrections may diverge lexically from the references while better preserving semantic accuracy.

\begin{table}[t]
\centering
\renewcommand{\arraystretch}{1}
\setlength{\tabcolsep}{0.8mm}
\begin{tabular}{lcccccc}
\toprule
& \multicolumn{4}{c}{\textbf{SARI-score}} & &\\
\cmidrule(lr){2-5}
\multicolumn{1}{c}{{\Large\textbf{Models}}} & \textbf{Keep} & \textbf{Add} & \textbf{Delete} & \textbf{Final} & \textbf{RG-2} & \textbf{GPT}\\
\midrule

\multicolumn{7}{l}{\small\textbf{Distantly Supervised Baselines}} \\
\quad LIFE\textsubscript{t5} & 71.82 & 26.30 & 85.94 & 61.35 & 0.63 & 0.45 \\
\quad VENCE\textsubscript{t5} & 63.53 & 5.98 & 84.01 & 51.17 & 0.58 & 0.20  \\
\midrule

\multicolumn{7}{l}{\small\textbf{Few-shot Baselines}} \\
\quad GPT\textsubscript{4o-8shot} & \underline{83.35} & \underline{44.69} & 87.09 & \underline{71.71} & 0.68 & \underline{0.84} \\
\quad LLaMA3-1b & 40.66 & 5.76 & 60.88 & 35.76 & 0.20 & 0.26 \\
\midrule

\multicolumn{7}{l}{\small\textbf{Do-Nothing Baseline}} \\
\quad Do-Nothing & 63.89 & 4.47 & 100.00 & 56.12 & 0.61 & 0.10 \\
\midrule

\multicolumn{7}{l}{\small\textbf{Our Methods}} \\
\quad CECoR\textsubscript{T5-sft} & 78.36 & 35.96 & \textbf{93.70} & 69.34 & \underline{0.70} & 0.52 \\
\quad CECoR\textsubscript{L3-1b-sft} & 76.09 & 31.19 & \underline{90.35} & 65.88 & 0.67 & 0.49 \\
\quad CECoR\textsubscript{L3-1b-rl} & 76.92 & 30.08 & 81.58 & 62.86 & 0.60 & 0.67 \\
\quad CECoR\textsubscript{L3-3b-sft} & \textbf{84.01} & \textbf{46.87} & 87.07 & \textbf{72.65} & \textbf{0.71} & 0.49 \\
\quad CECoR\textsubscript{L3-3b-rl} & 72.36 & 17.36 & 77.81 & 55.84 & 0.50 & \textbf{0.94} \\

\bottomrule
\end{tabular}
\caption{Comparison on the \textbf{FECDATA}.}
\label{tab:fecdata_results}
\end{table}

\textbf{In-domain Single-hop Correction.}
To examine CECoR’s effectiveness in simpler factual correction settings, we train our models using our method and evaluate them on the single-hop dataset FECDATA.
As shown in Table~\ref{tab:fecdata_results}, our models consistently outperform distantly supervised baselines and GPT-based systems across both rule-based metrics and LLM-as-a-judge evaluations.
Notably, CECoR\textsubscript{L3-3b-rl} achieves the highest GPT score, surpassing all baselines and its SFT counterpart, highlighting the benefit of reinforcement learning in aligning model outputs with human evaluation preferences.
While GPT-based models perform competitively due to their in-context learning capability, our smaller models deliver superior results with significantly fewer parameters and without additional task-specific tuning.
These results demonstrate that CECoR effectively handles single-hop factual correction when trained in-domain.

\begin{table}[h]
\centering
\renewcommand{\arraystretch}{1}
\setlength{\tabcolsep}{0.8mm}
\begin{tabular}{lcccccc}
\toprule
\cmidrule(lr){1-7}
& \multicolumn{4}{c}{\textbf{SARI-score}} & &\\
\cmidrule(lr){2-5}
\multicolumn{1}{c}{{\Large\textbf{Models}}} & \textbf{Keep} & \textbf{Add} & \textbf{Delete} & \textbf{Final} & \textbf{RG-2} & \textbf{GPT}\\

\midrule
\multicolumn{7}{l}{\small\textbf{Filters}} \\
\quad CECoR\textsubscript{T5-sft} & 91.18 & 49.35 & 89.76 & 76.76 & 0.87 & 0.43 \\
\quad CECoR\textsubscript{L3-1b-sft} & 89.94 & 43.27 & 85.30 & 72.83 & 0.84 & 0.34 \\
\quad CECoR\textsubscript{L3-1b-rl} & 87.59 & 41.22 & 71.05 & 66.62 & 0.81 & 0.62 \\

\midrule
\multicolumn{7}{l}{\small\textbf{Without Filters}} \\
\quad CECoR\textsubscript{T5-sft} & 90.93 & 47.66 & 89.52 & 76.04 & 0.87 & 0.41 \\
\quad CECoR\textsubscript{L3-1b-sft} & 88.47 & 41.70 & 82.83 & 71.00 & 0.82 & 0.25 \\
\quad CECoR\textsubscript{L3-1b-rl} & 84.46	& 36.14	& 64.65	& 61.75 & 0.77 & 0.48 \\

\bottomrule
\end{tabular}
\caption{Ablation study on the \textbf{HOVER} dataset.}
\label{tab:filters_results_hover}
\end{table}

\textbf{Effectiveness of Filtering.}
To assess the impact of our filtering pipeline, we conduct ablation experiments in the Tables~\ref{tab:filters_results_hover}. Results show consistent improvements across all models, especially in the SARI-Add and GPT-based scores, which capture the informativeness of edits and alignment with human preferences. Although the reinforcement learning stage operates on naturally occurring incorrect claims from the original dataset, the CECoR\textsubscript{L3-1b-rl} still exhibits notable performance gains. This indicates that filtered synthetic data used in the supervised fine-tuning phase provides a stronger initialization, thereby enabling more effective policy optimization in subsequent reinforcement learning. These findings underscore the importance of precise and reliable data construction for achieving robust factual correction performance.We also conduct ablation studies on additional multi-hop datasets;detailed results are provided in the Appendix 3.

\begin{table}[h]
\centering
\renewcommand{\arraystretch}{1}
\setlength{\tabcolsep}{0.8mm}
\begin{tabular}{lcccccc}
\toprule
\multicolumn{7}{c}{\textbf{Correction with Retrieved Evidence}} \\
\cmidrule(lr){1-7}
& \multicolumn{4}{c}{\textbf{SARI-score}} & &\\
\cmidrule(lr){2-5}
\multicolumn{1}{c}{{\Large\textbf{Models}}} & \textbf{Keep} & \textbf{Add} & \textbf{Delete} & \textbf{Final} & \textbf{RG-2} & \textbf{GPT}\\

\midrule
\multicolumn{7}{l}{\small\textbf{HOVER}} \\
\quad LIFE\textsubscript{t5} & 42.78 & 3.03 & 29.29 & 25.03 & 0.23 & 0.25 \\
\quad CECoR\textsubscript{T5-sft} & \textbf{90.05} & \textbf{37.50} & \textbf{87.87} & \textbf{71.81} & \textbf{0.86} & 0.30  \\
\quad CECoR\textsubscript{L3-1b-sft} & 89.07 & 37.41 & 80.94 & 69.14 & 0.82 & 0.34 \\
\quad CECoR\textsubscript{L3-1b-rl} & 85.34	& 33.41	& 61.68 & 60.14	& 0.78 & \textbf{0.45} \\
\midrule
\multicolumn{7}{l}{\textbf{FEVEROUS}} \\
\quad LIFE\textsubscript{t5} & 44.31 & 2.96 & 26.84 & 24.71 & 0.26 & 0.23 \\
\quad CECoR\textsubscript{T5-sft} & 89.02 & 25.09 & \textbf{89.14} & \textbf{67.75} & \textbf{0.84} & 0.33 \\
\quad CECoR\textsubscript{L3-1b-sft} & \textbf{90.10} & 23.75 & 76.17 & 63.34 & 0.63 & 0.30  \\
\quad CECoR\textsubscript{L3-1b-rl} & 87.90 & \textbf{25.83} & 62.89 & 58.87	& 0.72 & \textbf{0.48} \\
\midrule
\multicolumn{7}{l}{\textbf{FECDATA}} \\
\quad LIFE\textsubscript{t5} & 67.66 & 17.31 & 83.88 & 56.28 & 0.60 & 0.36 \\
\quad CECoR\textsubscript{T5-sft} & 74.13 & \textbf{26.67} & \textbf{91.35} & \textbf{64.05} & \textbf{0.66} & 0.42 \\
\quad CECoR\textsubscript{L3-1b-sft} & \textbf{74.14} & 24.51 & 86.21 & 61.62 & 0.63 & 0.47 \\
\quad CECoR\textsubscript{L3-1b-rl} & 71.78	& 19.15	& 75.14 & 55.36	& 0.48 & \textbf{0.78}\\

\bottomrule
\end{tabular}
\caption{Experimental results from training and testing with retrieved evidence on HOVER, FEVEROUS and FECDATA.}
\label{tab:retrieve_evidence}
\end{table}

\textbf{Robustness to Retrieved Evidence.}
To evaluate how our method performs with imperfect evidence, we conduct additional experiments where both training and testing use retrieved rather than gold passages. For HOVER, we use the October 2017 Wikipedia dump processed by Yang et al.~\cite{yang2018hotpotqadatasetdiverseexplainable}, which includes the introductory sections of 5.2 million articles. For retrieval, we adopt BM25~\cite{INR-019}, implemented via Pyserini~\cite{lin2021pyserinieasytousepythontoolkit}, and select the top-3 retrieved paragraphs as evidence. As shown in Table~\ref{tab:retrieve_evidence}, using retrieved evidence introduces noise and leads to a moderate performance drop compared to gold annotations. However, our models consistently surpass the LIFE baseline across all datasets. Notably, LIFE bypasses the filtering stage in this setting, since none of its synthetic examples generated using retrieved evidence are able to pass its filter. This further demonstrates that our framework not only handles imperfect evidence effectively, but also maintains strong correction performance under noisy and distant supervision, reflecting its practical potential in real-world applications.

\begin{table}[t]
\centering
\renewcommand{\arraystretch}{1}
\setlength{\tabcolsep}{0.8mm}
\begin{tabular}{lcccccc}
\toprule
\multicolumn{7}{c}{\textbf{HOVER-FECDATA}}  \\
\cmidrule(lr){1-7} 
& \multicolumn{4}{c}{\textbf{SARI-score}} & & \\
\cmidrule(lr){2-5}
\multicolumn{1}{c}{{\Large\textbf{Models}}} & \textbf{Keep} & \textbf{Add} & \textbf{Delete} & \textbf{Final} & \textbf{RG-2} & \textbf{GPT} \\
\midrule

\multicolumn{7}{l}{\small\textbf{Distantly Supervised Baselines}} \\
\quad LIFE\textsubscript{t5} & 59.70 & 5.20 & 70.15 & 45.01 & 0.52 & 0.26 \\
\quad VENCE\textsubscript{t5} & 62.17 & 5.03 & 80.79 & 49.33 & 0.56 & 0.16 \\

\midrule
\multicolumn{7}{l}{\small\textbf{Our Methods}} \\
\quad CECoR\textsubscript{T5-sft} & 72.29 & 24.06 & \textbf{93.76} & \textbf{63.37} & \textbf{0.64} & 0.41 \\
\quad CECoR\textsubscript{L3-1b-sft} & 72.56 & 22.34 & 90.92 & 61.94 & 0.61 & 0.41 \\
\quad CECoR\textsubscript{L3-1b-rl} & \textbf{73.35} & \textbf{24.49} & 85.93	& 61.26 & 0.60 & \textbf{0.51} \\
\bottomrule
\end{tabular}
\caption{Cross-domain evaluation: models trained on multi-hop datasets HOVER and tested on the single-hop dataset FECDATA. }
\label{tab:generalization_results}
\end{table}

\textbf{Cross-domain Transfer to Single-hop Tasks.}
We further assess cross-domain generalization by testing models trained exclusively on multi-hop datasets HOVER on the human-annotated single-hop dataset FECDATA.
As shown in Table~\ref{tab:generalization_results}, our models, particularly CECoR\textsubscript{T5-sft} and CECoR\textsubscript{L3-1b-rl}, significantly outperform distantly supervised baselines (LIFE and VENCE) across all metrics, despite never being trained on single-hop data.
In particular, CECoR\textsubscript{L3-1b-rl} achieves the highest GPT-based score on HOVER→FECDATA (0.51), indicating that reinforcement learning improves robustness even under domain shifts.
These findings confirm that our multi-hop training pipeline produces correction models with strong transferability to simpler, structurally distinct single-hop tasks without requiring additional adaptation.We further conduct cross-domain experiments on additional multi-hop datasets and detailed results are provided in the Appendix 4.

\begin{figure}[h]
  \centering
\includegraphics[width=1\linewidth]{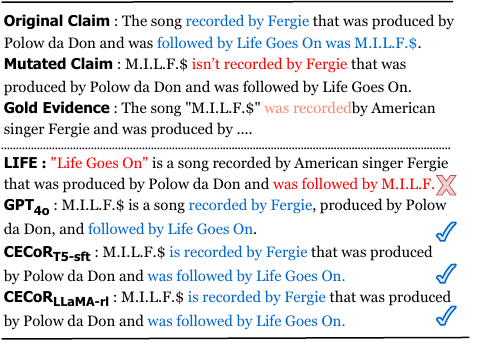}
\caption{Multi-hop factual correction.}
\label{fig:case_study1}
\end{figure}

\begin{figure}[h]
  \centering
\includegraphics[width=1\linewidth]{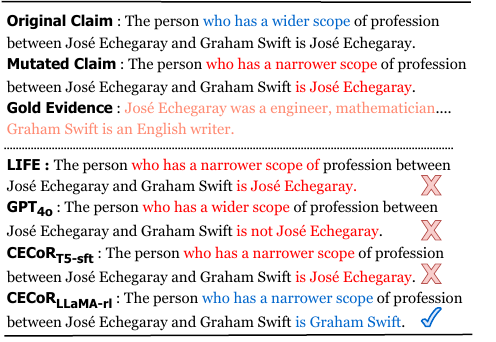}
\caption{Comparative reasoning via RL.}
\label{fig:case_study2}
\end{figure}

\textbf{Case Studies.}
To qualitatively evaluate our method’s capability in addressing complex multi-hop factual errors, we present two representative examples. These cases highlight the strengths of our RL-enhanced model and reveal the limitations of previous baselines and prompting-based LLMs. As shown in Figure~\ref{fig:case_study1}, all supervised variants of CECoR and GPT\textsubscript{4o} accurately correct the factual error, whereas the LIFE baseline produces a hallucinated output inconsistent with the evidence. This demonstrates CECoR’s ability to generate factually accurate and semantically coherent revisions in multi-hop scenarios. The second example (Figure~\ref{fig:case_study2}) involves implicit comparative reasoning. Only our RL-optimized model produces the correct correction, while other models either misinterpret the logical relation or leave the claim unmodified. This result underscores the effectiveness of reinforcement learning in aligning model outputs with complex semantic preferences beyond surface-level similarity. Overall, these qualitative findings corroborate our quantitative results: the RL-enhanced CECoR excels at resolving subtle, logically entangled errors that rule-based metrics often undervalue, yet are consistently preferred in human-aligned evaluations.

\section{Conclusion}
In this paper, we introduce CECoR, a reasoning-aware framework for factual error correction that addresses the limitations of prior methods under the atomic fact assumption. By decomposing multi-hop claims into structured logical chains and injecting semantically grounded errors at the reasoning-step level, CECoR generates high-fidelity training data that enables effective supervised learning. Reinforcement learning further enhances robustness by aligning model outputs with real-world factual inconsistencies.
Extensive experiments demonstrate that CECoR achieves state-of-the-art performance on complex multi-hop datasets and generalizes well to single-hop correction and noisy retrieved evidence, confirming its wide applicability.

\bibliographystyle{elsarticle-num}
\bibliography{kbs}

\end{document}

\begin{frontmatter}
\title{Supplementary Material}

\end{frontmatter}

\section*{Appendix}

\section{Statistics of Data Types}
\begin{table}[h]
\centering
\small
\renewcommand{\arraystretch}{1.15}
\setlength{\tabcolsep}{5pt}
\begin{tabular}{@{} l l c r r r @{}}
\toprule
\textbf{Dataset} & \textbf{Split} & \textbf{Hops} & \textbf{SUP} & \textbf{REF} & \textbf{Total} \\
\midrule

\multirow{8}{*}{HOVER}
& \multirow{4}{*}{Train} 
& 2     & 6496  & 2556  & 9052  \\
&       & 3     & 3271  & 2813  & 6084  \\
&       & 4     & 1256  & 1779  & 3035  \\
&       & Total & 11023 & 7148  & 18171 \\
\cmidrule(lr){2-6}
& \multirow{4}{*}{Val}
& 2     & 521   & 605   & 1126  \\
&       & 3     & 968   & 867   & 1835  \\
&       & 4     & 511   & 528   & 1039  \\
&       & Total & 2000  & 2000  & 4000  \\

\midrule
\multirow{2}{*}{FEVEROUS}
& Train & /     & 11616 & 15312 & 26928 \\
& Test  & /     & 1551  & 1411  & 2962  \\

\midrule
\multirow{2}{*}{FECDATA}
& Train & /     & 37961 & 20075 & 58036 \\
& Test  & /     & 1593  & 2289  & 3882  \\
\bottomrule
\end{tabular}
\caption{Dataset composition statistics. \textbf{SUP} and \textbf{REF} denote the numbers of instances labeled as \texttt{SUPPORTS} and \texttt{REFUTES}, respectively.}
\label{tab:status}
\end{table}

Table~\ref{tab:status} reports the dataset composition used in our experiments.
HOVER and FEVEROUS are multi-hop fact-checking benchmarks; however, only HOVER provides explicit hop-count annotations, so we further break down its Train/Val splits by hop number (2/3/4) and label type, where \textbf{SUP} and \textbf{REF} denote \texttt{SUPPORTS} and \texttt{REFUTES}, respectively.
In contrast, FEVEROUS does not include hop-count labels, and we therefore report split-level statistics only.
FECDATA is a single-hop factual error correction dataset, for which we also provide split-level label counts.
Overall, these statistics highlight differences in dataset scale and label distribution across benchmarks, which is important for interpreting correction performance and generalization results.

\section{CECoR Hyperparameters}
\subsection{SFT-LLaMA}
\begin{table}[htbp]
\centering
\renewcommand{\arraystretch}{1.12}
\setlength{\tabcolsep}{7pt}
\begin{tabular}{@{} l l @{}}
\toprule
\textbf{Hyperparameter} & \textbf{Value} \\
\midrule
Training Epochs & 1 \\
Per-device Batch Size & 4 \\
Gradient Accumulation Steps & 8 \\
Effective Batch Size & 32 \\
Initial Learning Rate & 2e-5 \\
Optimizer & AdamW \\
Mixed Precision & bfloat16 (bf16) \\
\bottomrule
\end{tabular}
\caption{Supervised fine-tuning (SFT) hyperparameters for LLaMA.}
\label{tab:sft-hyperparams}
\end{table}

Table~\ref{tab:sft-hyperparams} lists the hyperparameters used to fine-tune our LLaMA backbone under the SFT stage.
We use a per-device batch size of 4 with 8-step gradient accumulation, yielding an effective batch size of 32 updates per optimization step, which stabilizes training under limited GPU memory.
We train for a single epoch to reduce overfitting to synthetic supervision and to keep the comparison across settings consistent.
AdamW with a initial learning rate of $2\times10^{-5}$ is adopted as a widely used default for instruction-tuned LLM fine-tuning, while bf16 mixed precision improves throughput and memory efficiency without changing the underlying optimization objective.

\subsection{SFT-T5}

\begin{table}[htbp]
\centering
\renewcommand{\arraystretch}{1.12}
\setlength{\tabcolsep}{7pt}
\begin{tabular}{@{} l l @{}}
\toprule
\textbf{Hyperparameter} & \textbf{Value} \\
\midrule
Training Epochs & 1 \\
Per-device Batch Size & 4 \\
Gradient Accumulation Steps & 8 \\
Effective Batch Size & 32 \\
Initial Learning Rate & 2e-5 \\
Optimizer & AdamW \\
Mixed Precision & bfloat16 (bf16) \\
\bottomrule
\end{tabular}
\caption{Supervised fine-tuning (SFT) hyperparameters for t5-base.}
\label{tab:sft-t5-hyperparams}
\end{table}

Table~\ref{tab:sft-t5-hyperparams} reports the SFT configuration for \texttt{t5-base}.
We intentionally keep the core optimization settings aligned with those used for LLaMA (same effective batch size, learning rate, optimizer, and precision) to isolate the impact of model architecture rather than tuning choices.
The single-epoch schedule provides a controlled training budget and avoids excessive adaptation to the synthetic pairs, which helps maintain generalization when evaluating on datasets with different error patterns and evidence quality.

\subsection{The parameters of RL-LLaMA}
\begin{table}[htbp]
\centering
\renewcommand{\arraystretch}{1.12}
\setlength{\tabcolsep}{7pt}
\begin{tabular}{@{} l l @{}}
\toprule
\textbf{Hyperparameter} & \textbf{Value} \\
\midrule
Training Algorithm & GRPO \\
Per-device Batch Size & 4 \\
Gradient Accumulation Steps & 8 \\
Effective Batch Size & 32 \\
Initial Learning Rate & 2e-5 \\
Mixed Precision & bfloat16 (bf16) \\
Sampling Temperature & 0.7 \\
\bottomrule
\end{tabular}
\caption{Reinforcement learning (GRPO) hyperparameters for LLaMA.}
\label{tab:rl-llama-hyperparams}
\end{table}

Table~\ref{tab:rl-llama-hyperparams} summarizes the hyperparameters used in the reinforcement learning stage.
We adopt GRPO as the policy optimization algorithm and keep the optimizer, learning rate, precision, and effective batch size consistent with SFT to ensure that performance changes primarily reflect the training objective rather than optimization artifacts.
The sampling temperature is set to 0.7 to balance exploration and output stability during policy updates, encouraging diverse candidate corrections while maintaining fluency.
Overall, this configuration is designed to improve factuality and preference alignment on naturally occurring incorrect claims without introducing additional tuning degrees of freedom.

\section{Effectiveness of Filtering}
\begin{table}[htbp]
\centering
\renewcommand{\arraystretch}{1}
\setlength{\tabcolsep}{0.8mm}
\begin{tabular}{lcccccc}
\toprule
\multicolumn{7}{c}{\textbf{FEVEROUS}} \\
\cmidrule(lr){1-7}
& \multicolumn{4}{c}{\textbf{SARI-score}} & &\\
\cmidrule(lr){2-5}
\multicolumn{1}{c}{{\Large\textbf{Models}}} & \textbf{Keep} & \textbf{Add} & \textbf{Delete} & \textbf{Final} & \textbf{RG-2} & \textbf{GPT}\\

\midrule
\multicolumn{7}{l}{\small\textbf{Filters}} \\
\quad CECoR\textsubscript{T5-sft} & 94.39 & 54.18 & 89.48 & 79.35 & 0.90 & 0.67 \\
\quad CECoR\textsubscript{L3-1b-sft} & 92.40 & 45.31 & 81.42 & 73.04 & 0.84 & 0.62 \\
\quad CECoR\textsubscript{L3-1b-rl} & 84.88	& 33.56	& 59.06	& 59.17 & 0.73 & 0.74 \\

\midrule
\multicolumn{7}{l}{\small\textbf{Without Filters}} \\
\quad CECoR\textsubscript{T5-sft} & 94.27 & 54.55 & 89.23 & 79.34 & 0.89 & 0.66 \\
\quad CECoR\textsubscript{L3-1b-sft} & 90.90 & 40.05 & 77.45 & 69.47 & 0.79 & 0.61 \\
\quad CECoR\textsubscript{L3-1b-rl} & 69.57	& 11.58	& 75.89	& 52.35 & 0.36 & 0.65 \\

\bottomrule
\end{tabular}
\caption{Ablation study on the \textbf{FEVEROUS} dataset.}
\label{tab:filters_results_feverous}
\end{table}

Table~\ref{tab:filters_results_feverous} reports an ablation study on FEVEROUS to quantify the effect of our filtering pipeline during synthetic data construction. Overall, applying filters consistently improves model performance, with the most pronounced gains observed for CECoR\textsubscript{L3-1b-rl}: the GPT score increases from 0.65 to 0.74 and the SARI-Final score rises from 52.35 to 59.17 when filters are enabled. This pattern suggests that higher-quality synthetic pairs produced by filtering provide a stronger initialization for subsequent optimization, which is especially beneficial when reinforcement learning is applied on naturally occurring incorrect claims. These results support the claim that precise and reliable synthetic data construction is critical to robust factual error correction.

\section{Cross-domain Transfer to Single-hop Tasks.}
\begin{table}[htbp]
\centering
\renewcommand{\arraystretch}{1}
\setlength{\tabcolsep}{0.8mm}
\begin{tabular}{lcccccc}
\toprule
\multicolumn{7}{c}{\textbf{FEVEROUS-FECDATA}}  \\
\cmidrule(lr){1-7} 
& \multicolumn{4}{c}{\textbf{SARI-score}} & & \\
\cmidrule(lr){2-5}
\multicolumn{1}{c}{{\Large\textbf{Models}}} & \textbf{Keep} & \textbf{Add} & \textbf{Delete} & \textbf{Final} & \textbf{RG-2} & \textbf{GPT} \\
\midrule

\multicolumn{7}{l}{\small\textbf{Distantly Supervised Baselines}} \\
\quad LIFE\textsubscript{t5} & 65.18 & 12.37 & 76.34 & 51.30 & 0.51 & 0.45 \\
\quad VENCE\textsubscript{t5} & 63.00 & 5.43 & 81.73 & 50.06 & 0.57 & 0.18 \\

\midrule
\multicolumn{7}{l}{\small\textbf{Our Methods}} \\
\quad CECoR\textsubscript{T5-sft} & 69.96 & 18.83 & \textbf{91.03}	& 59.94 & 0.57 & 0.42 \\
\quad CECoR\textsubscript{L3-1b-sft} & 72.00 & \textbf{21.19} & 90.90 & \textbf{61.36} & \textbf{0.59} & 0.44 \\
\quad CECoR\textsubscript{L3-1b-rl} & \textbf{74.52} & 19.28 & 82.92 & 58.91 & 0.49 & \textbf{0.66} \\
\bottomrule
\end{tabular}
\caption{Cross-domain evaluation: models trained on multi-hop datasets FEVEROUS and tested on the single-hop dataset FECDATA. }
\label{tab:generalization_results}
\end{table}

Table~\ref{tab:generalization_results} presents cross-domain transfer results where models trained on the multi-hop dataset FEVEROUS are evaluated on the human-annotated single-hop benchmark FECDATA. Despite never being trained on single-hop correction pairs, all CECoR variants outperform the distantly supervised baselines LIFE and VENCE across key metrics, demonstrating strong transferability from multi-hop supervision to simpler single-hop corrections. Among our methods, CECoR\textsubscript{L3-1b-rl} achieves the best GPT-based score (0.66), indicating that reinforcement learning further improves robustness and preference alignment under domain shift. In contrast, the SFT variants obtain higher ROUGE-2 and stronger SARI components in some cases (e.g., CECoR\textsubscript{L3-1b-sft} attains the highest SARI-Final of 61.36 and RG-2 of 0.59), reflecting more conservative, reference-aligned editing behavior. Together, these results corroborate that our multi-hop training pipeline can produce correction models that generalize well to structurally different single-hop tasks without additional adaptation.

\section{LLM Prompting Configurations}
\subsection{Multi-hop Claim Decomposition}
As introduced in Section~2.3, CECoR relies on an explicit reasoning representation to enable step-level manipulation of multi-hop claims. We therefore prompt an LLM to decompose each multi-hop claim into a Python-like pseudocode program composed of three primitive operations—\texttt{Question}, \texttt{Verify}, and \texttt{Predict}. The prompt includes in-context exemplars that illustrate how to express multi-hop verification as a sequence of interpretable steps, and it explicitly constrains the model to output only the decomposed program without additional explanations. This controlled decomposition provides a structured interface between evidence-grounded sub-facts and editable reasoning units, serving as the foundation for subsequent reasoning-aware error injection.
\begin{tcolorbox}[
  title=Prompt for Multi-hop Claim Decomposition,
  colback=white,
  colframe=black,
  boxrule=0.8pt,
  arc=2mm,
  left=4pt,
  right=4pt,
  top=4pt,
  bottom=4pt,
  breakable
]

You are an expert in breaking down multi-hop questions into Python pseudocode.Generate a python-like program that describes the reasoning steps required to verify the claim step-by-step. You can call three functions in the program: 1. Question() to answer a question; 2. Verify() to verify a simple claim; 3. Predict() to predict the veracity label. Several examples are given as follows.\\

\# The claim is that Howard University Hospital and Providence Hospital are both located in Washington, D.C.
def program():
    fact\_1 = Verify("Howard University Hospital is located in Washington, D.C.")
    fact\_2 = Verify("Providence Hospital is located in Washington, D.C.")
    label = Predict(fact\_1 and fact\_2)\\

\# The claim is that WWE Super Tuesday took place at an arena that currently goes by the name TD Garden.
def program():
    answer\_1 = Question("Which arena the WWE Super Tuesday took place?")
    fact\_1 = Verify(f"{answer\_1} currently goes by the name TD Garden.")  
    label = Predict(fact\_1)
Next is the question, please note that only the split results should be output, without any reasoning.
The claim is that [[CLAIM]]
\end{tcolorbox}

\subsection{Reasoning-aware Error Injection}
To synthesize realistic incorrect claims without manual correction pairs in Section~2.4, we prompt an LLM to inject factual errors directly into the decomposed reasoning program. The prompt operationalizes our Decomposition-and-Injection paradigm by specifying module-wise transformations for \texttt{Predict}, \texttt{Verify}, and \texttt{Question}, producing multiple corrupted variants from a single correct claim. Importantly, the prompt enforces strict constraints on content preservation: each generated sentence must remain fluent while retaining all clauses implied by the original program, differing only through targeted, semantically plausible but incorrect substitutions or negations. The output is restricted to a JSON list of corrupted sentences, which facilitates large-scale automatic generation and downstream filtering.
\begin{tcolorbox}[
  title=Prompt for Reasoning-aware Error Injection,
  colback=white,
  colframe=black,
  boxrule=0.8pt,
  arc=2mm,
  left=4pt,
  right=4pt,
  top=4pt,
  bottom=4pt,
  breakable
]

You are a fact error injection expert. Your task is to take a correct multi-hop sentence that has been split into pseudo Python code and generate multiple incorrect multi-hop sentences by injecting similar but wrong entities or answers within each module. Please strictly follow the steps below and do not output anything other than the final list.\\
1. Process the Predict module  
   - Scan down to label = Predict(fact\_1 and fact\_2 …)  
   - If there are fact\_i more than two, extract all the shared entity/entities E that appear in them  
   - Replace all occurrences of E with a semantically similar but incorrect entity E', and rephrase all into a fluent sentence — call this fei\_1  
   - When rephrasing, do not omit any sentence. All content from the Python code must be combined. The rephrased sentence must include every clause from the original.\\
2. Process the Verify module
   - For each fact\_i = Verify("…"):
   - Inside each fact, identify a unique entity F (avoid using the shared E from above)
    a. Replace F with a similar but incorrect entity F', rephrase into a fluent sentence — call this fei\_2
    b. Convert the sentence into a negated version (i.e., make the fact false), still fluent — call this fei\_3  
   - When rephrasing, do not omit any sentence. All content from the Python code must be combined. The rephrased sentence must include every clause from the original.\\
3. Process the Question module  
   - Execute answer = Question("Q") to get the original answer A 
   - Replace A with a similar but wrong answer B
   - Construct a new question Q' that is very similar in form to Q but the correct answer is now B
   - Rephrase into a fluent sentence combining Q' and B — call this fei\_4  
   - When rephrasing, do not omit any sentence. All content from the Python code must be combined. The rephrased sentence must include every clause from the original.\\
4. Output Format  
- Output **only** the list of sentences in this format (do not use markdown, code blocks, or explanations):  
[
  "…fei\_1…",
  "…fei\_2…",
  …
]\\
- Do not include any reasoning, explanation, or extra formatting (e.g., no ```python or \textbackslash n).
- Ensure this is a valid JSON array of strings.
- Note: When rephrasing, do not omit any sentence. All content from the Python code must be combined. The rephrased sentence must include every clause from the original.
- Each sentence must retain all unaltered and alter information from the original sentence without omission.

Here are some examples: 

Oirginal:The writer of Day of the Iguana worked as a comedian and not Norman Foster.
Python:```python\textbackslash n def program():\textbackslash n    fact\_1 = Verify(\textbackslash"The writer of Day of the Iguana worked as a comedian.\")\textbackslash n    fact\_2 = Verify(\textbackslash"Norman Foster did not work as a comedian.\textbackslash")\textbackslash n    label = Predict(fact\_1 and fact\_2)\textbackslash n```
Output:["The writer of Day of the Iguana worked as a producer and not Norman Foster.","The writer of Day of the Iguana worked as a comedian and not Orson Welles.","The writer of Day of the Iguana worked as a comedian and so did Norman Foster.","The writer of Diary of a Wimpy Kid worked as a comedian and not Norman Foster.","Neither the writer of Day of the Iguana nor Norman Foster worked as a comedian."]

Oirginal:Owen Wilson acted in a film with Tom Wu. He also voiced Lightning McQueen in the \textbackslash"Cars\textbackslash" franchise.
Python:```python\textbackslash n def program():\textbackslash n    answer\_1 = Question(\textbackslash"Which film did Owen Wilson act in with Tom Wu?\textbackslash")\textbackslash n    fact\_1 = Verify(f\textbackslash"Owen Wilson acted in the film {answer\_1} with Tom Wu.\textbackslash")\textbackslash n    fact\_2 = Verify(\textbackslash"Owen Wilson voiced Lightning McQueen in the 'Cars' franchise.\textbackslash")\textbackslash n    label = Predict(fact\_1 and fact\_2)\textbackslash n```
Output:["Luke Wilson acted in a film with Tom Wu. He also voiced Lightning McQueen in the \textbackslash"Cars\textbackslash" franchise.","Owen Wilson acted in a film with Tom Wu. He also voiced Mater in the \textbackslash"Cars\textbackslash" franchise.","Owen Wilson acted in a film with Tom Wu. He didn’t voice Lightning McQueen in the \textbackslash"Cars\textbackslash" franchise.","Owen Wilson didn’t act in a film with Tom Wu. He also voiced Lightning McQueen in the \textbackslash"Cars\textbackslash" franchise.","Owen Wilson acted in a film with Jet Li. He also voiced Lightning McQueen in the \textbackslash"Cars\textbackslash" franchise."]

Here is the question:\\
Oirginal:[[claim]]\\
Python:[[prediction]]\\
Output:
\end{tcolorbox}

\subsection{Filtering}
As synthetic corruption can occasionally introduce ungrammatical outputs or fail to flip factuality, we apply an LLM-assisted filtering pipeline to ensure data reliability before supervised fine-tuning. We use two complementary prompts: (i) a hop-count estimator that predicts the number of reasoning hops implied by a sentence, which helps retain multi-hop candidates and filter out degenerate single-hop or trivial cases; and (ii) a quality validator that checks whether a corrupted sentence is fluent, grammatically correct, and contains an injected factual error. Both prompts enforce minimal outputs (a hop number or a binary Yes/No decision) to reduce ambiguity and improve consistency. Together, they provide lightweight but effective quality control for constructing high-fidelity synthetic training pairs.
\begin{tcolorbox}[
  title=Prompt for determine whether it is a multi-hop claim,
  colback=white,
  colframe=black,
  boxrule=0.8pt,
  arc=2mm,
  left=4pt,
  right=4pt,
  top=4pt,
  bottom=4pt,
  breakable
]

You are an expert in determining multi-hop sentences. You will be given a sentence, and your task is to determine how many hops (pieces of reasoning) the sentence involves. Only output the final number — do not include any extra explanation or content. Below are a few examples:\\
    Question:Skagen Painter Peder Severin Krøyer favored naturalism along with Theodor Esbern Philipsen and the artist Ossian Elgström studied with in the early 1900s.",\\
    Answer:3\\
    Question:Red, White \& Crüe and Mike Tyson both fight.\\
    Answer:2\\
    Question:WWE Super Tuesday took place at an arena that currently goes by the name TD Garden.",\\
    Answer:2\\
    Here is the question:\\
    Question:[[claim]]\\
    Answer:
\end{tcolorbox}

\begin{tcolorbox}[
  title=Prompt for grammatically correct and errors exist,
  colback=white,
  colframe=black,
  boxrule=0.8pt,
  arc=2mm,
  left=4pt,
  right=4pt,
  top=4pt,
  bottom=4pt,
  breakable
]

You are an expert in assessing the quality of sentences after factual error injection. You will be given a sentence that has undergone factual error injection, and you need to judge its quality. Requirements:

1.The sentence is semantically coherent.
2.The sentence is grammatically correct.
3.The factual error has been successfully injected, i.e., the sentence contains an incorrect fact.

If all three conditions are satisfied, output "Yes". Otherwise, output "No".

Note: Output only “Yes” or “No” and do not include any additional content.

Sentence after factual error injection: [[claim]]

Output:
\end{tcolorbox}

\subsection{LLM as a judge}
To mitigate the limitations of rule-based automatic metrics, we adopt an LLM-as-a-judge protocol for semantic evaluation, consistent with recent findings on LLM-based assessment. The judge prompt scores a candidate correction conditioned on the original erroneous claim and the supporting evidence, explicitly balancing three criteria: fluency, factual correctness with respect to evidence, and semantic similarity/preservation of key content from the original. This evaluation setup is also aligned with our reinforcement learning stage.The output is constrained to a single scalar score in $[0,1]$ with three-decimal precision, enabling stable aggregation and model comparison. 
\begin{tcolorbox}[
  title=Prompt for judge,
  colback=white,
  colframe=black,
  boxrule=0.8pt,
  arc=2mm,
  left=4pt,
  right=4pt,
  top=4pt,
  bottom=4pt,
  breakable
]

You are a strict language quality evaluator. You will be given an original erroneous sentence, a piece of supporting evidence, and a corrected sentence based on that evidence. You must evaluate the quality of the correction based on the original erroneous sentence and the supporting evidence. The score should range from 0 to 1, and the final result must be a single decimal number (rounded to three decimal places) representing the overall quality of the corrected sentence.

Evaluation Criteria:

Fluency: Is the sentence grammatically correct and natural-sounding?
Correctness: Does the correction accurately address factual, logical, or grammatical errors in the original sentence? Does it align with the provided evidence and correctly amend any misleading or incorrect information?Does it cover and correct **all** necessary content from the original sentence?
Similarity: Does the corrected sentence preserve the original meaning without unnecessary changes?Does it maintain **all essential factual points** from the original? Has any important information from the original been omitted or lost?

Instructions:

Output only one line containing a single decimal number between 0 and 1 (inclusive), rounded to three decimal places.
Do not include any reasoning, explanation, labels, or additional output — only the number.
Be strict: only assign a perfect score (1.000) if the corrected sentence is fully fluent, logically and factually correct, and perfectly retains the intended meaning **with all key elements present**.
If the corrected sentence is identical to the original and the original contains factual or logical errors, assign a low score (e.g., 0.0–0.2).
If the corrected sentence contains placeholder content (e.g., "answer here", "reasoning process here") or is meaningless, assign a score of 0.000.
If the corrected sentence fails to fix a factual error, omits key information, or introduces an incorrect fact, penalize the score accordingly.

Now, rate the following:

Mutated: [[mutated]]

Evidence: [[evidence]]

Corrected: [[output\_text]]
\end{tcolorbox}

\bibliographystyle{elsarticle-num}
\bibliography{kbs}